\documentclass[conference]{IEEEtran}   
\IEEEoverridecommandlockouts
\usepackage{cite}
\usepackage{amsmath,amssymb,amsfonts}
\usepackage{algorithmic}
\usepackage{graphicx}
\usepackage{textcomp}
\usepackage{xcolor}
\usepackage{hyperref}
\def\BibTeX{{\rm B\kern-.05em{\sc i\kern-.025em b}\kern-.08em
    T\kern-.1667em\lower.7ex\hbox{E}\kern-.125emX}}
    
\begin{document}

\title{\Large{Design Space Exploration and Explanation via Conditional Variational Autoencoders in Meta-model-based Conceptual Design of Pedestrian Bridges}\\
\thanks{*Corresponding author. Stefano-Franscini-Platz 5, 8093 Zürich, Switzerland, \textit{E-Mail adress:} kraus@ibk.baug.ethz.ch\\
$^{1}$ both authors contributed equally to the presented manuscript.\\
The authors would like to thankfully acknowledge the facilities of Design++ at ETH Zürich and the funding through ETH Foundation grant No. 2020-HS-388 (provided by Kollbrunner/Rodio) as well as the SDSC Project "Domain-Aware AI-augmented Design of Bridges (DAAAD Bridges)".}}


\author{\IEEEauthorblockN{Vera M. Balmer$^{a,1}$, Sophia V. Kuhn$^{a,1}$, Rafael Bischof$^{b}$, Luis Salamanca$^{b}$, Walter Kaufmann$^{a}$,\\ Fernando Perez-Cruz$^{b}$, Michael A. Kraus$^{a,*}$}
\vspace{10pt}
\IEEEauthorblockA{
$^{a}$ \small{ETH Zurich, Chair of Concrete Structures and Bridge Design and Design++, Stefano-Franscini-Platz 5, 8093 Zürich, Switzerland}\\
$^{b}$ Swiss Data Science Center (SDSC), Turnerstr. 5, 8006 Zürich, Switzerland
}}

\maketitle
\pagestyle{plain}
\begin{abstract}

For conceptual design, engineers rely on conventional iterative (often manual) techniques. Emerging parametric models facilitate design space exploration based on quantifiable performance metrics, yet remain time-consuming and computationally expensive. Pure optimisation methods, however, ignore qualitative aspects (e.g. aesthetics or construction methods). This paper provides a performance-driven design exploration framework to augment the human designer through a Conditional Variational Autoencoder (CVAE), which serves as forward performance predictor for given design features as well as an inverse design feature predictor conditioned on a set of performance requests. The CVAE is trained on 18’000 synthetically generated instances of a pedestrian bridge in Switzerland. Sensitivity analysis is employed for explainability and informing designers about (i) relations of the model between features and/or performances and (ii) structural improvements under user-defined objectives. A case study proved our framework's potential to serve as a future co-pilot for conceptual design studies of pedestrian bridges and beyond.

\end{abstract}


\section{Introduction}
The conceptual design phase in current architectural, engineering and construction (AEC) practice is generally disconnected from later phases of the building process, leading to an insufficient consideration of ultimate functionality, performance, material impacts, manufacturing and construction processes of the built structures. Yet, this phase is where the most influential design decisions for a building project are made, which greatly influence the material consumption and environmental impacts. In light of increasing urbanization and associated environmental concerns, these topics can no longer be ignored. In addition, digital workflows for design and planning as well as systematic creation of design databases are not common practice in AEC industry. Today, the initial design is mainly based on the prior knowledge of (i) the design team, and (ii) the manual analysis of a few similar reference projects. The practice of post-processing a relatively fixated design from the conceptual phase requires significant manual - and hence lengthy - labour with numerous translations, detailing, and re-iterations of the design models between early-stage designers, engineers and contractors involved in later stages \cite{kuhn2022ntab0}.

Recent breakthroughs in artificial intelligence (AI) and especially machine (ML) and deep learning (DL) have already had a transformative impact on many fields, including medicine, physics and finance. However, the application of these technologies in AEC is still in its infancy, due to the aforementioned domain-specific siloed thinking, non-digitized workflows and resulting processes and tools. This is also why traditional ML and DL approaches are insufficient for direct application to most AEC design problems due to lack of data and/or inconsistencies and biases on it. 

The computational design paradigm offers designers systematic and digital means of exploring vast parametrically generated design spaces. High-dimensional design spaces with a large number of design parameters are appealing, as these can potentially contain high-performing, yet unimagined or unexpected design solutions. However, they come with the curse of dimensionality and are hard to process for human designers, hence underlining the need for tools to assist in their exploration and account for human intuition. We propose to employ DL algorithms, specifically a variant of Conditional Variational Autoencoders (CVAE), to form the basis for an AI-based design co-pilot augmenting human creativity with the computational power of modern algorithms to support early design decision-making processes, turning them more informed, integrated, productive, and flexible.

\begin{figure}[ht]
    \centering
    \includegraphics[scale = 1]{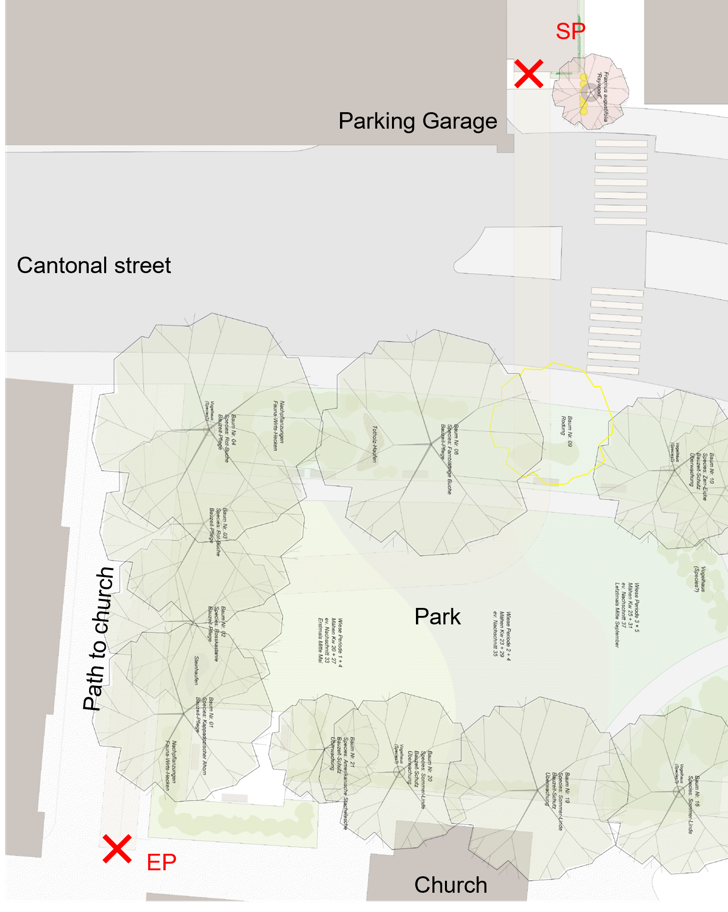}
    \caption[Content]{Project site, M 1:750, from Woodtli et al. \cite{woodtli_svec_2021}}
    \label{fig:ProjectSite}
\end{figure}

The objective of this research is to (i) develop a synthetic dataset generation pipeline for quantifyable (structural) design performances, (ii) to implement, train and evaluate a suitable design meta-model for the forward and inverse design setting, and (iii) to create a demonstrator application of an explainable and intuitive co-pilot for conceptual design. We demonstrate the applicability of this approach for the generic example of a bridge design project in St.~Gallen, Switzerland, cf. Fig.~\ref{fig:ProjectSite}. By integrating our ML algorithms into established Building-Information-Modelling (BIM) software (here Autodesk Revit 2022 \cite{autodeskrevit}), the proposed framework can easily be transferred to future bridge project scenarios (and beyond, e.g. to housing or office buildings) as soon as the parametric bridge model together with requested boundary conditions and performance criteria are defined within Revit. Therefore, our solution integrates seamlessly into the current design paradigm known by AEC domain experts. In combination with implemented explainable AI methods such as the Sensitivity Analysis (SA), this ubiquitousness ensures acceptance amongst engineers or project managers with limited expertise in ML or AI and fosters wide-spread use of such tools in practice. Further information can be found on the project \href{https://mkrausai.github.io/research/01_SciML/01_BH_PedestrianBridge_XAI/}{homepage}.

\section{Related work}

In recent years, computer vision has been the avant-garde when it comes to the development of deep generative models, such as Variational Autoencoders (VAE) \cite{vae_kingma} or Generative Adversarial Networks (GAN) \cite{good_fellow_gan}. In engineering applications, especially in civil engineering, data-driven design generation is a highly under-explored research area, mainly due to the lack of systematised, freely accessible datasets. However, generative models have recently been found to be powerful tools in generative modelling of automobiles \cite{burnap2016estimating}, mechanical components \cite{gan_wheels_design_oh}, performance-based shading device design for office buildings \cite{ercan2015performance} or floorplans \cite{house_gan_nauata, chaillou2020archigan}. In order for such tools to prevail in practice, they must provide an interface for including human experts into the design loop. 
Jang et al. \cite{Jang2022Generative} provide a definition of generative design as a computational design method for automated conduction of design exploration under consideration of user-defined constraints. Salamanca et al. \cite{salamanca2022augmented_semiramis} developed a conditional Autoencoder for finding optimal designs of a vertical garden under the objectives of rain occlusion or exposure to sunlight. Danhaive et al. \cite{Danhaive2021Design} proposed to utilise the latent space of a performance-conditioned VAE to provide an intuitive tool for structural design-space exploration of a spatial steel truss for the objective of structural mass. 

Yet, the cited approaches stayed far below AEC practice complexity when it comes to software integration as well as dimensionality of the design features and objectives in addition to neglecting the consideration of state-of-the-art structural design analysis requirements.

\begin{figure*}[ht]
    \centering
    \includegraphics[width=.8\linewidth]{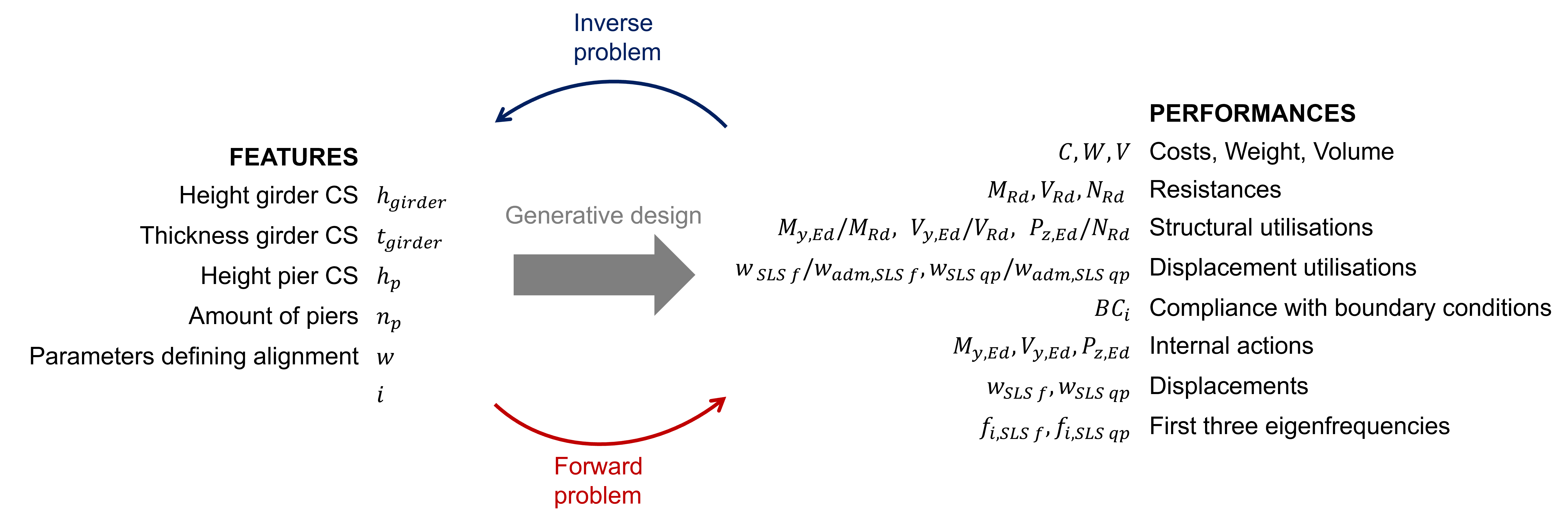}
    \caption{Overview of design features and performances for the bridge design task for synthetic data generation}
    \label{fig:DataSetParameters}
\end{figure*}

\section{Methods} \label{sec:Methods}
This section presents the methodology, workflow and software used for developing the design space learning and browsing framework for the pedestrian bridge design example.

\subsection{Bridge Project Description} \label{sec:bridgeDesignDescription}
The specific use-case used as demonstrator project for the proposed framework is a pedestrian bridge in St. Gallen, Switzerland, which is currently in the conceptual design phase. The pedestrian bridge shall provide a pedestrian crossing from an exit next to a parking garage (start point / SP in Fig.~\ref{fig:ProjectSite}), over a busy cantonal street and then through the St. Mangen park, to terminate next to a church (end point / EP in Fig.~\ref{fig:ProjectSite}). The currently existing surroundings of the bridge project as well as the corresponding coordinate system is given as a scanned point cloud within a BIM model in Autodesk Revit \cite{autodeskrevit}. A \href{https://polybox.ethz.ch/index.php/s/JAaBViw4TsBnJGq}{rendered walk-over the bridge} supports imagination.

The city of St. Gallen defined fixed boundary conditions for the project, cf. Fig.~\ref{fig:ProjectSite} and \cite{woodtli_svec_2021}: (i) SP and EP of the bridge are fixed to specific locations together with the request of straight alignments, (ii) a prescribed clearance height over the cantonal street and the path in the park need to be respected, (iii) to minimise cut down of trees due to cultural heritage protection, and (iv) a fixed bridge width of $b=2.50~m$.

\subsection{Design \{Representation, Features, and Performances\}}
The bridge design model is represented by a $d$-dimensional (Euclidean) vector of design feature variables $\mathbf{x} \in \mathcal{D} \subset \mathbb{R}^d$, which together with predefined feature intervals $\times_j [x_{j,min}, x_{j,max}], (j = 1,...,d)$ span the domain of the design space $\mathcal{D}$. Evaluating a performance metric $\mathcal{P}(\cdot)$ for a design may be computationally expensive and is a priori unknown except when computed for specific instances $\mathbf{y} = \mathcal{P}(\mathbf{x})$. For the design process, two situations have to be distinguished:
\begin{enumerate}
    \item \textit{Forward Design}: for the full set of design features $\mathbf{x} \in \mathcal{D}$ (e.g. cross-section parameters, alignment geometry,~etc.) we request the associated performance metrics $\mathcal{P}(\mathbf{x})$ (e.g. costs, structural utilisation,~etc.).
    \item \textit{Inverse Design}: for given performance metrics $y = \mathcal{P}(\mathbf{x})$ (i.e. costs, structural utilisation,~etc.) and potentially a subset of design features $\mathbf{\Tilde{x}} \in \mathbb{R}^{d_1}$ we request feasible (remaining) design features $\mathbf{x} \in \mathbb{R}^{d_2}$ with $d_1 + d_2 =d$.
\end{enumerate}
This study develops AI-based meta models for both design situations together with respective interaction interfaces, cf. Sec.~\ref{sec:data} and Sec.~\ref{sec:Results}.

In order to choose the dimensionality of the bridge design features, we adopt the propositions of SIA 112 \cite{SIA}, which specifies a level of development (LOD, \cite{abualdenien2021Ausarbeitungsgrade}) 300 for the task at hand. This implies certain parameters of the bridge project to be fixated and hence not subject to further design variation such as the bridge use (pedestrian bridge), the bridge typology (girder bridge) and the boundary conditions stated in Sec.~\ref{sec:bridgeDesignDescription}. The design features $x_j$ are provided on the left hand of Fig.~\ref{fig:DataSetParameters} and define the girder height $h_{girder}$ and thickness $t_{girder}$, the amount $n_{p}$ and dimensions $h_{p}$ of the piers as well as parameters $w,i$ of the NURBS curve for defining the bridge alignment.

\begin{table}[ht]
     \centering
     \caption{Design Performance Objectives $\mathcal{P}(\mathbf{x})$}
     \vspace{5pt}
     \label{tab:GoalParameters}
     \begin{tabular}{l|l|l}
          Performance & Considered & Simulator\\
          \hline
          Structural Utilisation &  &\\
          - Ultimate Limit State & yes &  Sofistik (FEA) \\
          - Serviceability Limit State & yes &  Sofistik (FEA)\\
          Structural Dynamics & yes & Sofistik (FEA) \\
          Costs & yes & Revit Dynamo\\
          Aesthetics & no & - \\
          Sustainability & no & - \\
     \end{tabular}
\end{table}

The set of performance metrics $y = \mathcal{P}(\mathbf{x})$ for the design of the pedestrian bridge is provided on the right hand of Fig.~\ref{fig:DataSetParameters}, where mostly structural safety and serviceability (resistances, utilisations, internal actions and displacements) criteria are employed next to structural dynamics aspects (eigenfrequencies), costs (computed via the weight and volume) and the boundary conditions for SP, EP and the trees as defined in Sec.~\ref{sec:bridgeDesignDescription}. The structural performance objectives together with the load definitions for the pedestrian bridge (such as loads due to dead weight, pedestrians, wind, snow and temperature) are derived from currently applicable design standards in Switzerland by SIA 261 and 262 \cite{SIA}. 

\subsection{Synthetic Data Generation Pipeline and Dataset} \label{sec:data}
Training deep latent generative models requires a significant amount of data which, for this study, is collected in a two-stage approach. At first, a central Latin Hypercube Sampling (LHS, \cite{mckay1979Comparison}) of the design space $\mathcal{D}$ is launched within the specifications as provided in Tab.~\ref{tab:InputsSampler}.
\begin{table}[ht]
    \centering
    \caption{Input definition for LHS}
    \label{tab:InputsSampler}
    \vspace{5pt}
    \begin{tabular}{l|c}
         Input & Value \\
         \hline
         Design Features $\mathbf{x}$ & \sffamily{[$h_{girder}, t_{girder}, n_{p}, h_{p}, i, w$]}\\
         Minimum bounds & \sffamily{[0.25, 0.1, 2, 0.5, 0, 0.01]} \\
         Maximum bounds & \sffamily{[2.5, 0.23, 8, 1.5, 4, 7.00]}  \\
         Amount of samples & \sffamily{18'000} \\
    \end{tabular}
\end{table}
In the second stage, the sampled design features $\mathbf{x}$ are handed to performance simulators to obtain the performance metrics $\mathcal{P}(\mathbf{x})$ as defined on the right hand side of Fig.~\ref{fig:DataSetParameters}. For obtaining design performances, either analytical expressions (for costs, weight, volume as well as compliance with boundary conditions) or the Finite-Element-Analysis (FEA) software "Sofistik" (remaining performances) is utilised. A parametric template for the FEA-based structural analysis was developed and connected to Revit via zero-touch nodes inside Dynamo for the FEA performance simulator to be able to evaluate the vast amounts of parameter samples in a standardised way. The parametric FEA template contains load definitions, a parametrized geometry description and evaluations for the structural and dynamic performances. The results are saved to text files and subsequently merged with the design features to form a data frame as basis for ML modelling. 

\subsection{Machine Learning Model}
\label{sec:model}
\begin{figure}[h]
    \centering
    \includegraphics[width=\linewidth]{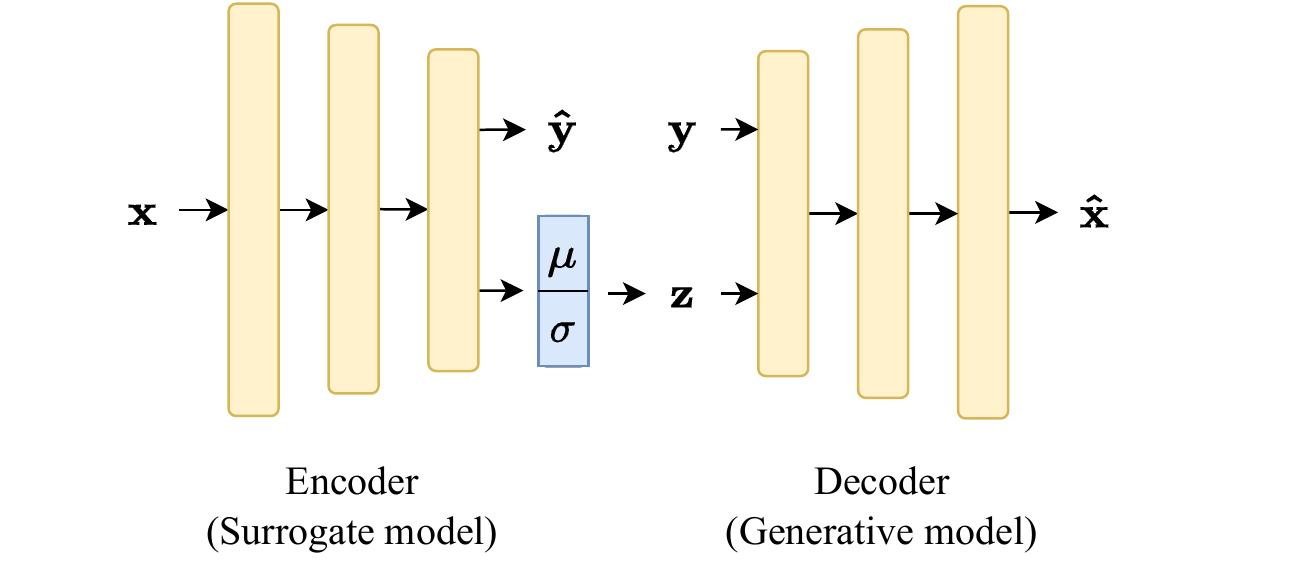}
    \caption{Architecture of our CVAE network acting both as surrogate as well as generative model.}
    \label{fig:cvae_architecture}
\end{figure}

The ML model used in this study is a variation of Conditional Variational Autoencoders (CVAE) \cite{cvae_sohn}. In light of having to solve both a forward as well as an inverse problem, we forgo feeding the conditional $\mathbf{y}$ to the encoder and instead let it predict the performance metrics together with a latent vector in two separate heads (cf. Fig~\ref{fig:cvae_architecture}). 

More specifically, the encoder receives the design features $\mathbf{x}$ as input and produces the outputs $\mathbf{\hat{y}}$ and $\mathbf{z}$, denoting the performance metrics and an arbitrary latent vector, respectively. Since the performance metrics may include categorical features with discrete values, we replace the predicted $\mathbf{\hat{y}}$ with the ground truth $\mathbf{y}$ before passing it to the decoder to avoid skewing the model towards continuous values which it is not going to encounter at inference time. Finally, $\mathbf{y}$ and $\mathbf{z}$ are passed through the decoder to produce $\mathbf{\hat{x}}$. Both the encoder and the decoder consist of an arbitrary number $l$ of MLP-Blocks, where each block is comprised of a fully-connected layer with leaky-ReLU activation and batch normalisation. 

We chose this architecture over a GAN because it is more robust to train and easier to thoroughly evaluate, as there is no straight-forward way of quantitatively estimating the "goodness" of a GAN. Furthermore, our framing allows to train a surrogate model (encoder) as well as a conditional generative model (decoder) simultaneously and is therefore tailored to the setting where both a forward and an inverse problem have to be solved. Once trained, the decoder generates new samples $\mathbf{\hat{x}}$ conditioned on some specified performance metrics $\mathbf{y}$. On the other hand, the encoder can be used to compute the mapping from $\mathbf{\hat{x}}$ to an approximate performance metric $\mathbf{\hat{y}}$, thus acting as a reliability estimator for designs generated by allowing to examine the reconstruction error between requested performance $\mathbf{{y}}$ and the predicted performance of the generated designs $\mathbf{\hat{y}}$.

\subsection{CVAE Objective Function}
In order to deter the encoder from including information about $\mathbf{y}$ into the latent vector $\mathbf{z}$, we decorrelate the two vectors by regularising for the covariance between $\mathbf{y}$ and $\mathbf{z}$ over the batch dimension \cite{covariance_regularisation_zbontar}:
\begin{align}
    \operatorname{Cov}[\mathbf{y},\mathbf{z}]=\mathrm{E}\left[(\mathbf{y}-\mathrm{E}[\mathbf{y}])^{\top}\mathbf{z}\right]
\end{align}
\begin{align}
    \mathcal{L}_{\textit{cov}}(\mathbf{y}, \mathbf{z}) = \frac{1}{d_y d_z}\sum_{i, j} \operatorname{Cov}[\mathbf{y},\mathbf{z}]_{i, j}^2
    \label{eq:decorrelation_term}
\end{align}
where $\mathbf{y} \in \mathbb{R}^{d_y}$ and $\mathbf{z} \in \mathbb{R}^{d_z}$. This term enforces the independence $p(\mathbf{y}, \mathbf{z}) = p(\mathbf{y})p(\mathbf{z})$ and avoids having to estimate the conditional $p(\mathbf{z \vert y})$ at generation time. Note that we can abstain from deducting $\mathrm{E}[\mathbf{z}]$ from $\mathbf{z}$ as its mean is already driven towards zero by the KL-divergence.

The total objective function of the CVAE is therefore comprised of the reconstruction loss $\mathcal{L}_{\textit{des}}$, the accuracy of the predicted performance metrics $\mathcal{L}_{\textit{perf}}$, the KL divergence and the decorrelation between $\mathbf{y}$ and $\mathbf{z}$:
\begin{align}
\label{eq:objective_function}
    \mathcal{L} = \lambda_1 \mathcal{L}_{\textit{des}}(\mathbf{x}, \mathbf{\hat{x}}) + \lambda_2 \mathcal{L}_{\textit{perf}}(\mathbf{y}, \mathbf{\hat{y}})\\
    + \lambda_3 \mathcal{L}_{\textit{KL}}(\mathbf{z}) + \lambda_4 \mathcal{L}_{\textit{cov}}(\mathbf{y}, \mathbf{z}) \nonumber
\end{align}
where $\lambda_i$ are hyperparameters for selecting the contribution of each term towards the total loss. Both $\mathcal{L}_{\textit{des}}$ and $\mathcal{L}_{\textit{perf}}$ consist of the mean squared error (MSE) loss for continuous features, or the crossentropy (CE) loss for discrete and categorical features, respectively.

\subsection{Explainability through Design Sensitivity Analysis}
Sensitivity analysis is a well-known method for performing topology optimisation by taking the derivative of the performance metrics in the output w.r.t. the design variables in the input of the model \cite{sensitivity_analysis_topology_optimization}. However, computing these derivatives in established solvers, e.g. leveraging the FEM, is expensive as they have to be estimated numerically. With neural networks being fully differentiable, this feature is readily implemented in our CVAE and can be computed very efficiently through Automatic Differentiation (AD). 

When inspecting the sensitivity of a performance metric $\frac{\partial \mathbf{\hat{y}_i}}{\partial \mathbf{x}}$ for a certain design $\mathbf{x}$, the designer receives information in which direction the design variables should be changed in order to improve the particular performance attribute. Furthermore, the distribution of sensitivities over a large set of designs yields information about the network's decision-making. An expert designer with prior knowledge in the design, analysis and construction of bridges can therefore, to a certain extent, estimate the model's reliability based on the relations it found between the features.

\section{Results \& Discussion} \label{sec:Results}
We sampled 18'000 instances of the pedestrian bridge together with their performances within the generative design as described in Sec.~\ref{sec:Methods} to form the dataset for subsequent CVAE training. While LHS of the design spaces $\mathcal{D}$ was conducted in a few seconds, obtaining performances of a batch of 600 design instances via FEA took on average around 55 minutes. 

\begin{figure*}[ht]
    \centering
    \includegraphics[width=0.44\linewidth]{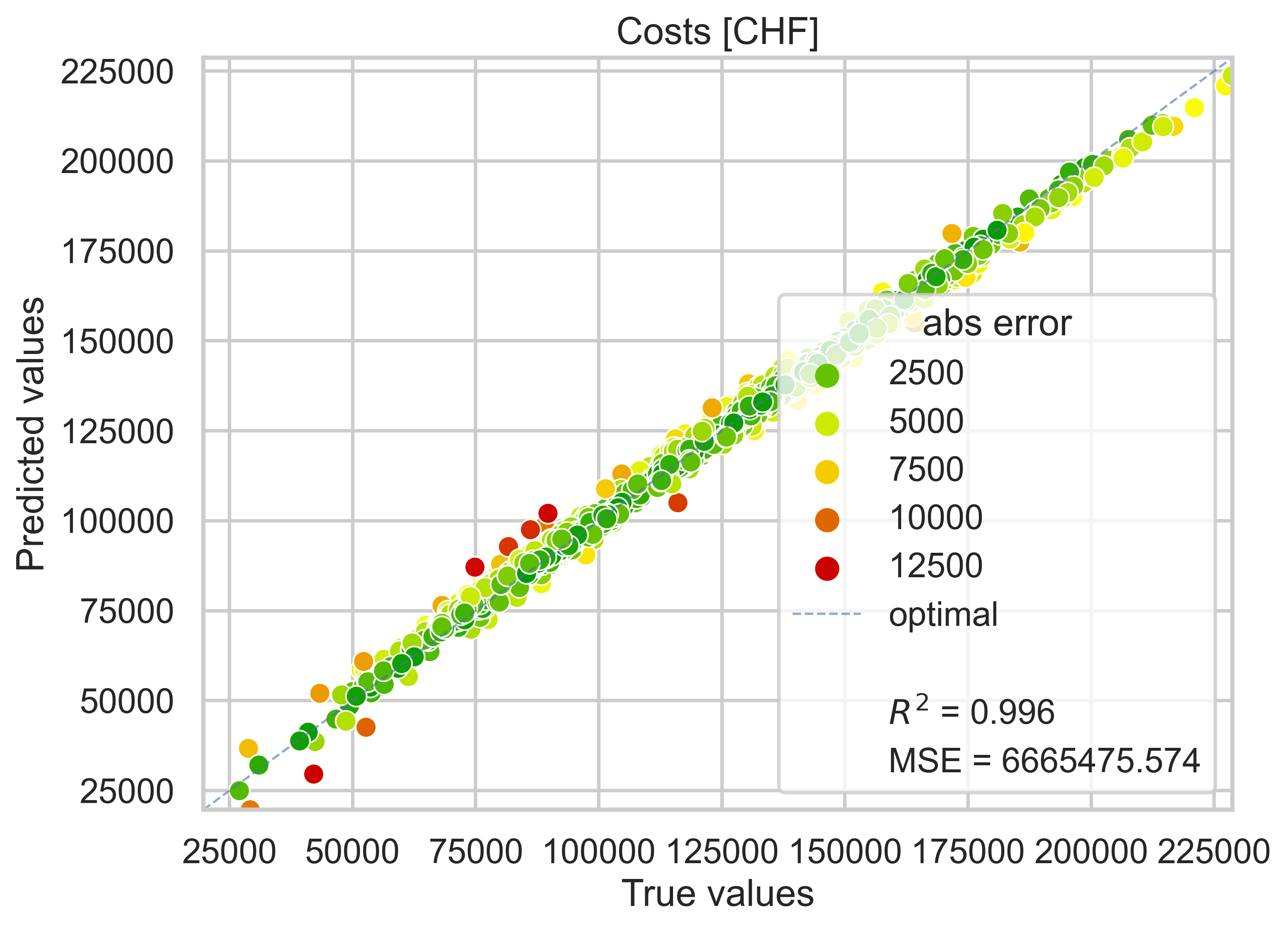}
    \hspace{2mm}
    \includegraphics[width=0.43\linewidth]{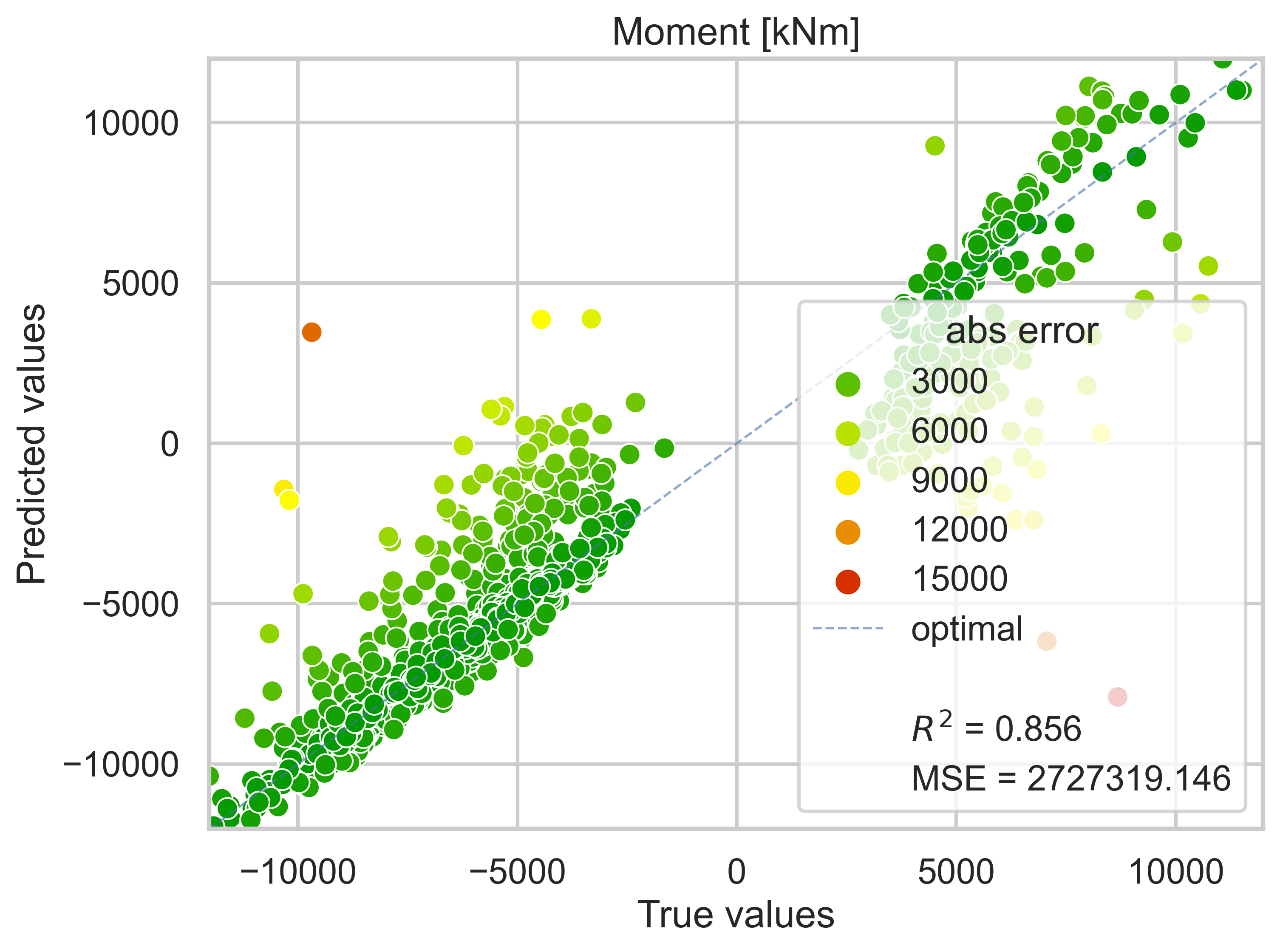}
    \includegraphics[width=0.42\linewidth]{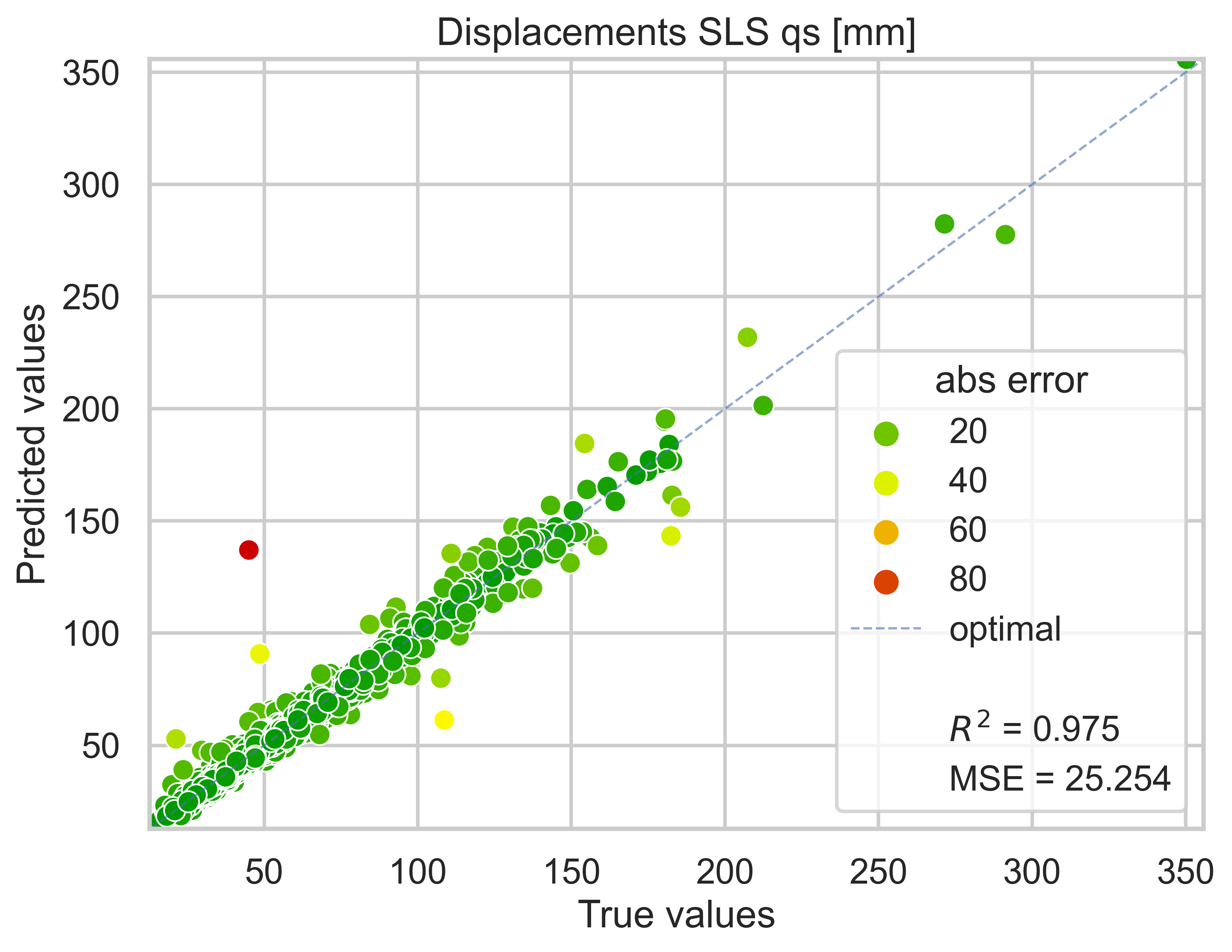}
    \hspace{5mm}
    \includegraphics[width=0.413\linewidth]{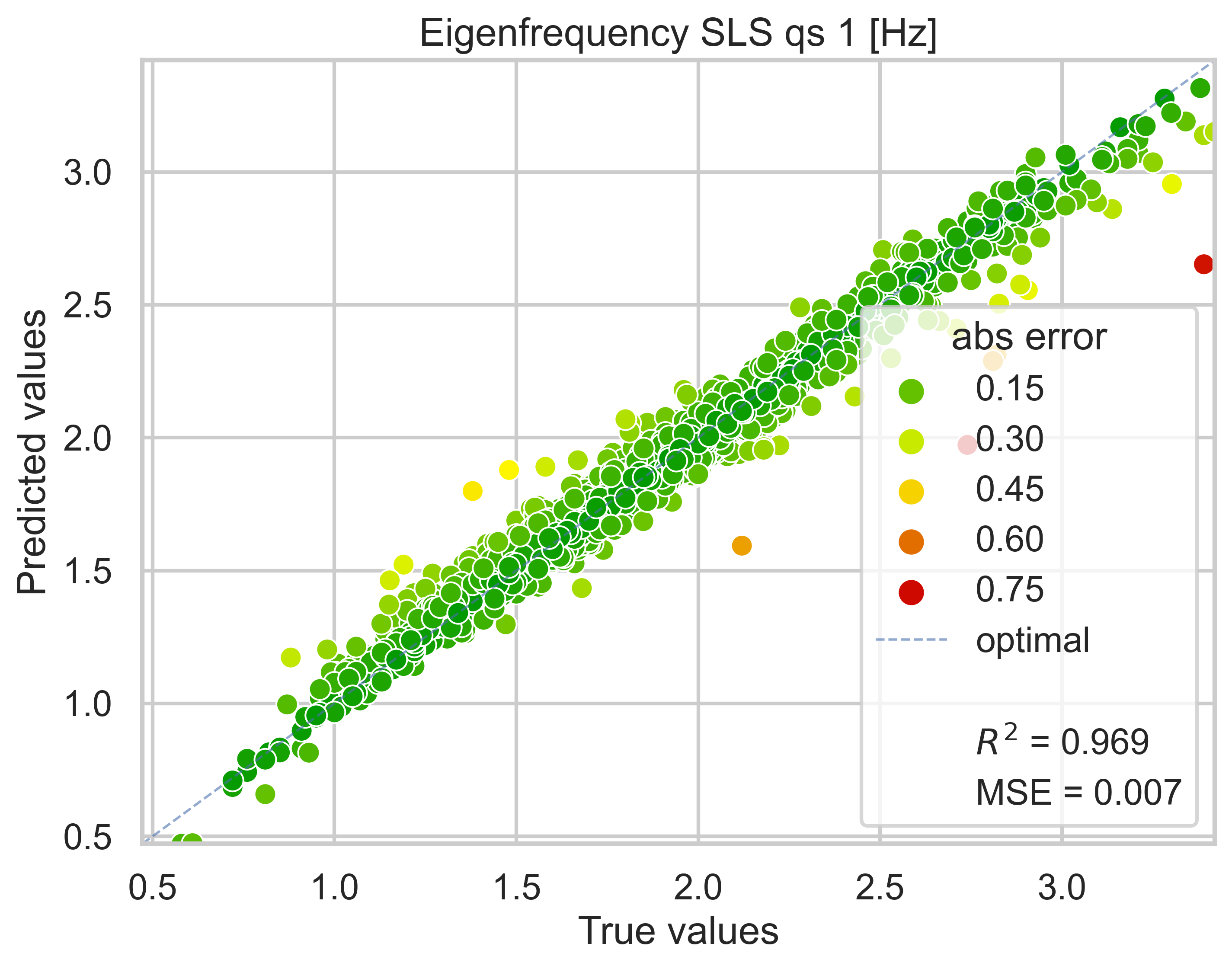}
    \caption{Model performance on four different performance metrics predicted by the encoder. Optimally, all points should lie on the diagonal.}
    \label{fig:results_diagonal_plots}
\end{figure*}

\begin{figure*}[ht]
    \centering
    \includegraphics[width=0.475\linewidth]{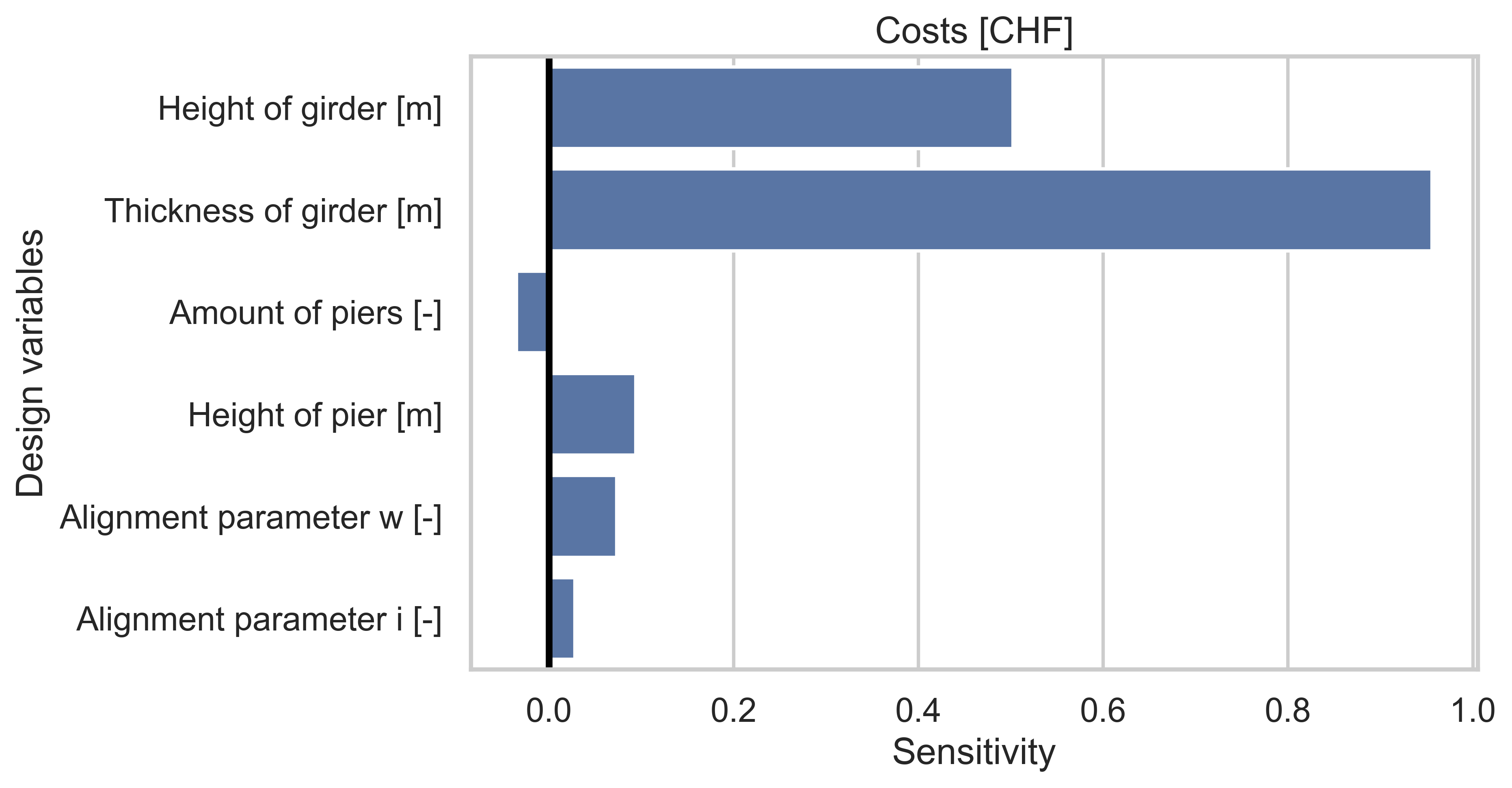}
    \hspace{2mm}
    \includegraphics[width=0.475\linewidth]{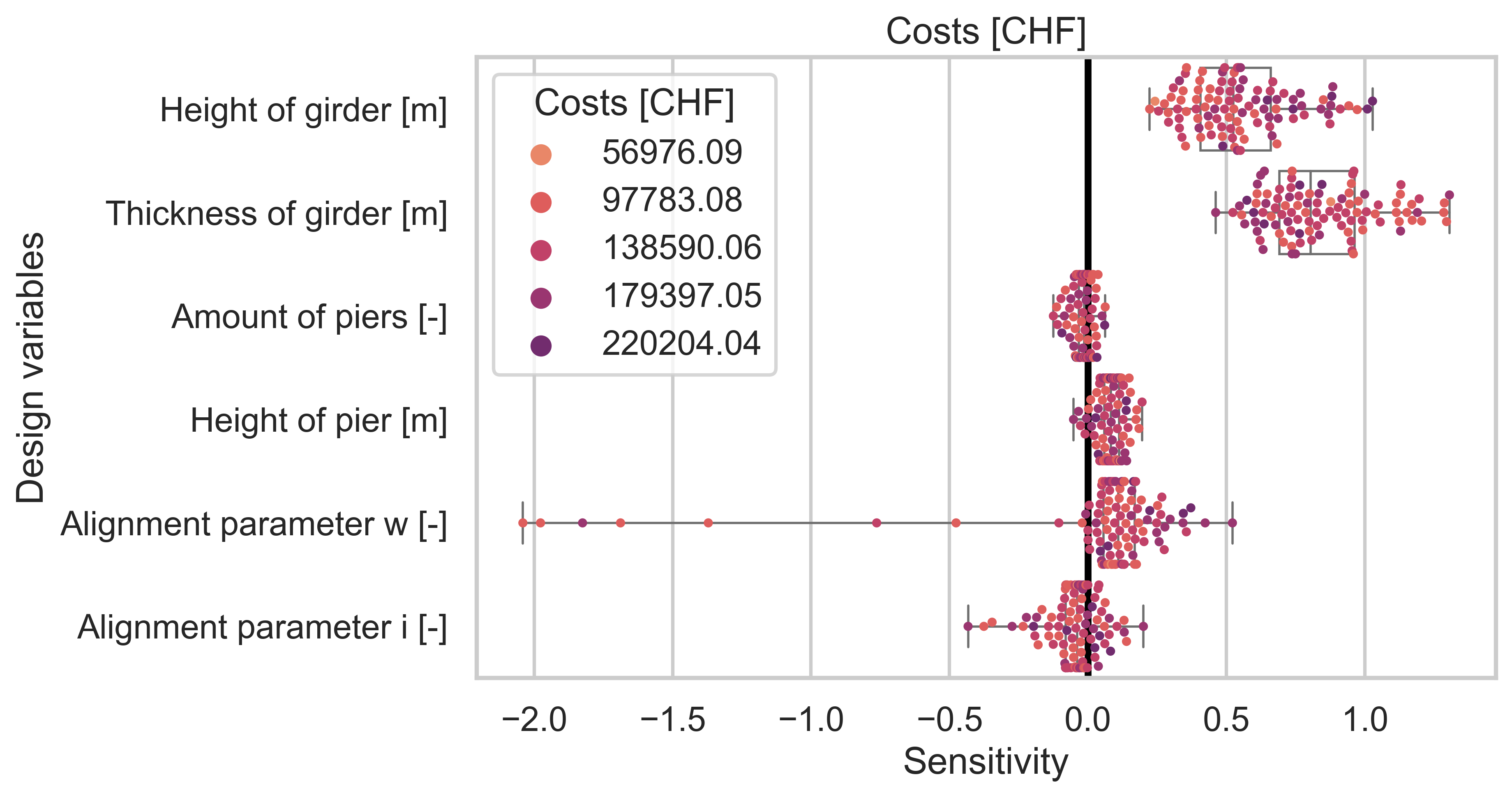}
    \caption{Sensitivities of the costs w.r.t. the input features (design variables) for: (left) a single user-chosen bridge, (right) a batch of 100 randomly generated bridges around the user-chosen bridge.}
    \label{fig:sensitivities_single_swarm}
\end{figure*}

For training the model and assessing its ability to generalise, the dataset was split into training, validation and test sets, where the training set accounted for 70\%, the validation set for 10\% and the test set for 20\% of the total number of samples. All continuous and ordinal features, both in $\mathbf{x}$ as well as in $\mathbf{y}$, were standardised to have mean $\mu = 0$ and a standard deviation $\sigma = 1$, while categorical features were one-hot encoded. Training was then performed using the Adam optimiser \cite{adam_optimiser_Kingma2015} with an initial learning rate $\eta = 0.001$ and a decrease of $\eta$ by a factor of 0.1 after every six epochs without improvement on the validation set. Finally, an early stopping was employed after 12 epochs without improvement.

\begin{figure}[h]
    \centering
    \includegraphics[width=0.8\linewidth]{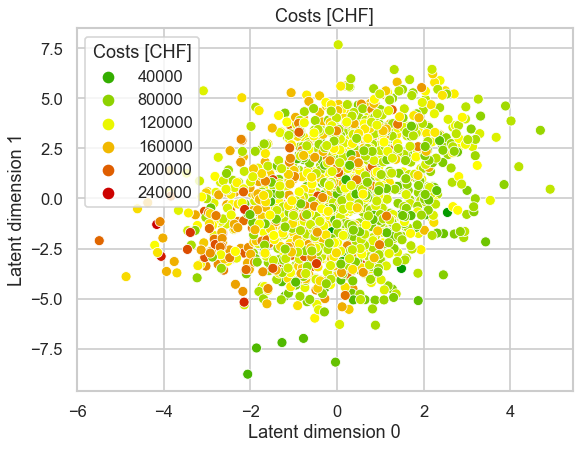} \\
    \includegraphics[width=0.8\linewidth]{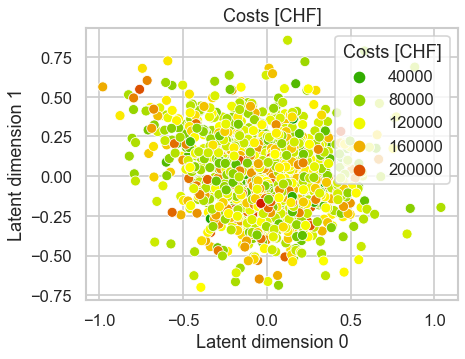}
    \vspace{2mm}
    \caption{Visualisation of the latent space with the points colored according to their respective cost without decorrelation (left) and with decorrelation (right).}
    \label{fig:latentspacevisualisation}
\end{figure}

The best-performing widths per MLP-Block were empirically found to be [128, 256, 512, 256, 128] in both the encoder and decoder, which accumulates to a sizeable amount of parameters and indicates that the complexity of this problem is non-negligible. The dimensionality of the latent space spanning $\mathbf{z}$ is set to 2 to allow for human interpretation. Finally, we set the weights of the loss terms to $\lambda_1 = 1$, $\lambda_2 = 10$, $\lambda_3 = 0.1$ and $\lambda_4 = 0.01$. Note that the effects of $\lambda_3$ for the KL-divergence and $\lambda_4$ for the decorrelation can be assessed by visually inspecting the latent space $\mathbf{z}$ and coloring the points according to their value of $\mathbf{y}$ (cf. Fig.~\ref{fig:latentspacevisualisation}). If $\mathbf{z}$ is not distributed normally, the hyperparameter $\lambda_3$ should be increased, whereas $\lambda_4$ should be increased if the scatter plot reveals correlations between $\mathbf{z}$ and $\mathbf{y}$. 

\begin{figure*}[h]
    \centering
    \includegraphics[width=\linewidth]{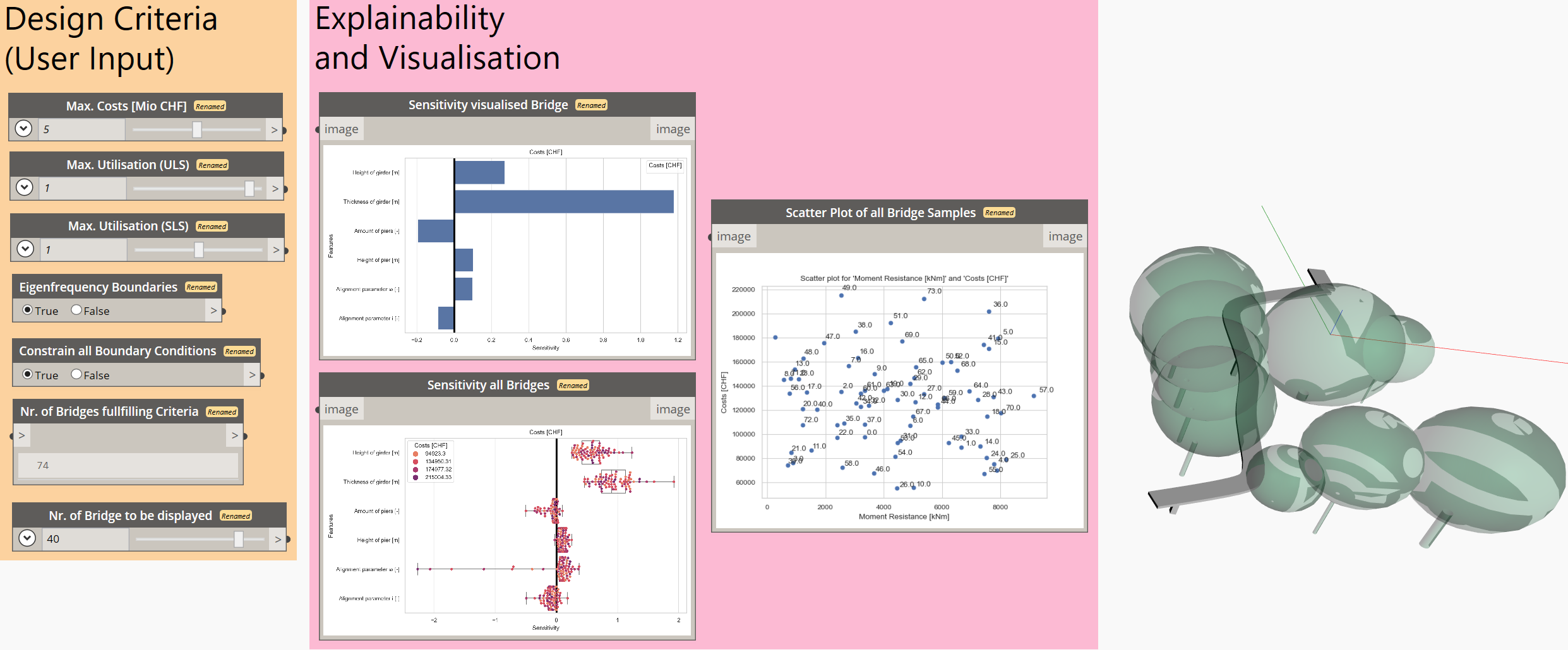}
    \caption{User interface for interaction with the meta-model for the inverse bridge design setting (Screenshot Dynamo)}
    \label{fig:InverseGUI}
\end{figure*}




Given the brevity of this paper, we show only selected results. First, in Fig.~\ref{fig:results_diagonal_plots} we show the performance on the test set of the surrogate model, i.e. the forward mapping from design features $\mathbf{x}$ to performances $\mathbf{y} = \mathcal{P}(\mathbf{x})$, on a selection of four features. It can be observed that the main bulk of predictions lies accurately on the diagonal and hence proves that the CVAE predicted every target well on average while the prediction's standard deviation is dependent on the respective target quantity. In addition, the prediction intervals are reasonable for a realistic pedestrian bridge design.

Fig.~\ref{fig:sensitivities_single_swarm} (right) depicts the sensitivities of the predicted costs w.r.t. the design variables for 100 generated bridges in a swarm plot. In addition to the horizontal placement of the bridges based on the calculated sensitivities, we also coloured the points according to their cost. This can give further insight, as cheaper bridges may be influenced by different design variables compared to more costly structures. Our trained model found a clear, positive relation between the height and thickness of the girder and the final costs of the bridge. This is obviously expected, and helps validating the sensitivity analysis, as the costs in this synthetic dataset were calculated based on the amount of material necessary for building the bridge. Both the height and the thickness of the girder have a direct influence on the amount of concrete used in the structure. The alignment parameters define the curvature and thus the length of the bridge and have therefore an effect on the costs as well, though to a lesser extent.

Fig.~\ref{fig:sensitivities_single_swarm} (left) shows the derivative of the costs w.r.t. the design variables for a single generated bridge. This plot is useful to designers for making informed decisions on how to improve the bridge according to the user-chosen objectives. 
%

Figure~\ref{fig:InverseGUI} shows the prototype of a user interface for the inverse design situation developed within Revit Dynamo. It provides the user with sliders and check-boxes, allowing to set desired performance metrics such as ranges for costs or the maximum structural utilisation in the ultimate as well as serviceability limit state. More fine grained requests for different objectives, as well as an additional visualisation of the latent space and the mapping of the objectives, is also possible, yet has been omitted for the sake of clearness. The right hand side of the user interface displays the sensitivity plots as well as a scatter plot for finding the Pareto front. Finally, a rendering of the generated bridge is shown, which can be inspected either on the screen in 2D, or in 3D through virtual reality on smartphones or dedicated devices.

We demonstrated our developments to a selected group of researchers and practitioners (15 persons) within a hands-on session. The collected feedback towards ergonomics, efficiency and quality prove our framework to be intuitive, efficient, reliable and to bear a great potential for applications in practice.

\section{Conclusions and Future Research}
Subspace learning establishes a new paradigm for performance-conditioned exploration of design spaces, which is neither an optimisation setting nor a random process. Rather, it provides an intuitive and efficient cartography of the vastness of these design spaces. Instead of replacing human intuition with predefined, deterministic, quantitative rules, the AI acts as a design collaborator/co-pilot that augments the human designer's intuition on the problem at hand.

This research provides a variation of CVAEs tailored to forward and inverse design situations. We showed the potential of our CVAE in meta-modelling (i) the forward problem by providing a surrogate to estimate more efficiently and quickly design performances given design features, (ii) compression of complex design spaces into continuous, smooth, low-dimensional design subspaces. With a forward pass through our CVAE being extremely efficient, it can provide performance conditioned designs in quasi real-time and thus augment human designers by providing instant feedback and proposals during the iterative prototyping phase. Furthermore, with analytical derivatives inherently provided in neural networks, we demonstrated that the sensitivity analysis serves as powerful tool for both design optimisation as well as model interpretability. The latter is crucial for building trust and achieving wide acceptance of this kind of design augmentation tools in the AEC domain. The collected user responses prove our framework possesses the potential to find wide application in industry and research as a co-pilot for conceptual design studies in the AEC domain beyond pedestrian bridges. 

Future research will be concerned with multiple open points identified within this research. First, a more efficient performance-based sampling of the design space to create a meaningful synthetic dataset under multiple design objectives has to be investigated. Finally, more effort is necessary for improving the human-machine interfaces. The reported AI model can be enhanced with a recommender system that allows users to directly interact with the results and thus capture less quantifiable metrics like aesthetics. In addition, displaying results to the human designer and collecting feedback could be shifted from a computer screen to immersive environments employing knowledge from education in civil engineering with extended reality \cite{kraus2021struct,kraus2022mixed}.

\bibliographystyle{./IEEEtran}
\bibliography{./IEEEabrv,./IEEEexample,./IEEEfull}

\end{document}